# An AI-assisted workflow for object detection and data collection from archaeological catalogues

**Authors**

Kevin Klein[1,2]*, Antoine Muller[3], Alyssa Wohde[4], Alexander V. Gorelik[1,2], Volker Heyd[5], Ralf Lämmel[6], Yoan Diekmann[1], Maxime Brami[1,2]*

**Affiliation**

[1]Palaeogenetics Group, Institute of Organismic and Molecular Evolution (iomE), Johannes Gutenberg University Mainz, Mainz, Germany.

[2]Vor- Und Frühgeschichtliche Archäologie, Institut Für Altertumswissenschaften, Johannes Gutenberg University Mainz, Mainz, Germany.

[3]SFF Centre for Early Sapiens Behaviour (SapienCE), University of Bergen, Bergen, Norway.

[4]University of Koblenz, Koblenz, Germany

[5]Department of Cultures / Archaeology, University of Helsinki, Helsinki, Finland

[6]Software Languages Team, University of Koblenz, Koblenz, Germany

* Correspondence to Kevin Klein (kkevin@students.uni-mainz.de) and Maxime Brami (mbrami@uni-mainz.de)

# Abstract

Reconciling the ever-increasing volume of new archaeological data with the abundant corpus of legacy data is fundamental to making robust archaeological interpretations. Yet, combining new and existing results is hampered by inconsistent standards in the recording and illustration of archaeological features and artefacts. Attempts at collating data from images in existing publications first involve scouring the substantial body of existing literature, followed by extracting images that require onerous manual preprocessing steps, like re-scaling, re-orienting, and re-formatting. While the sample sizes of such manual analyses are curtailed by these problems, recent developments in AI and big data methods are poised to accelerate and automate large syntheses of existing data.

This paper introduces an AI-assisted workflow capable of creating uniform archaeological datasets from heterogeneous published resources. The associated software (*AutArch*) takes large and unsorted PDF files as input, and uses neural networks to conduct image processing, object detection, and classification. Objects commonly found in archaeological catalogues – like graves, skeletons, ceramics, ornaments, stone tools, and maps – are reliably detected. Accompanying elements of the illustrations, like North arrows and scales, are automatically used for orientation and scaling. Outlines are then extracted with contour detection, allowing whole-outline morphometrics. Detected objects, contours, and other automatically retrieved data can be manually validated and adjusted via *AutArch*'s graphical user interface.

While we test this workflow on third millennium BCE Central European graves and Final Neolithic/Early Bronze Age arrowheads from Northwest Europe, this method can be applied to the vast number of artefacts and archaeological features for which shape, size, and orientation holds technological, functional, cultural, and/or temporal significance. This AI-assisted workflow has the potential to speed-up, automate, and standardise data collection throughout the discipline, allowing more objective interpretations and freeing sample sizes from budget and time constraints.

# Highlights

- Automating archaeological data collection using legacy resources
- Fast and standardised recording for big data archaeology
- Contour detection and encoding allowing formal analyses of shapes

# Graphical Abstract

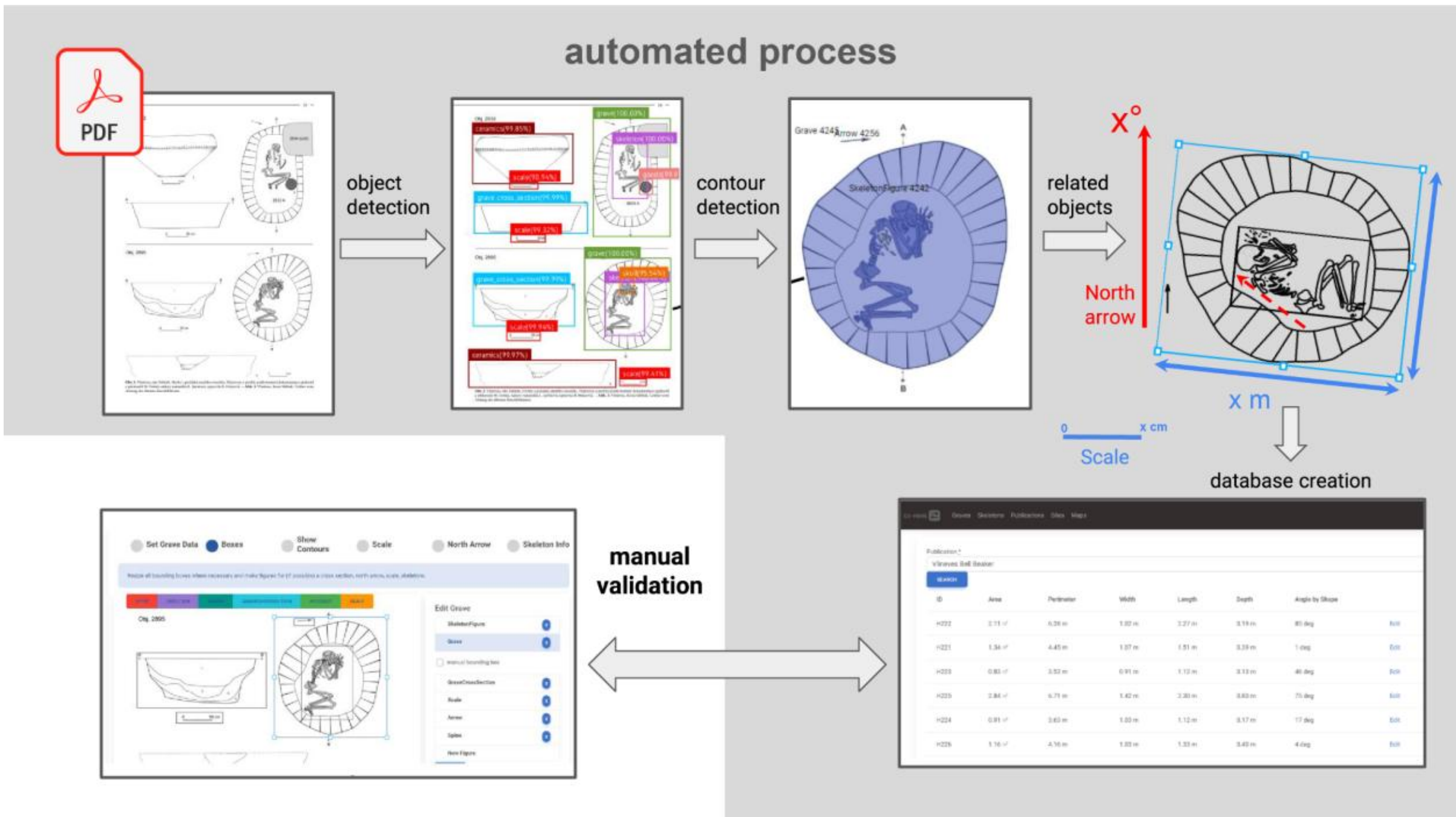

*Graphical Abstract. Overview of the AI-assisted workflow (here: Corded Ware site of Vliněves* [1]*, Czech Republic).*

# Video

A video demonstrating the key aspects of the AI-assisted workflow, as supported by the *AutArch* software, can be downloaded at this address: https://seafile.rlp.net/f/50ce9cc0771f4afc9cd0/

# 1 Introduction

## 1.1 Context: AI and Big Data in Archaeology

The rate of data collection in archaeology is outpacing our ability to meaningfully interpret this data. Each new excavation accrues more archaeological evidence than can be meaningfully analysed, and the existing body of data grows with each passing year. Continual methodological improvements to archaeological analyses also necessitate re-analysing legacy datasets to make our present results comparable with those from previous decades. Together, these factors make holistically interpreting existing and new archaeological data an insurmountable task with present methods. This poses the most urgent challenge to archaeologists, whose primary goal is preserving *and* interpreting information about the past. Recent advances in AI present opportunities for applying big data methods to this otherwise overwhelming corpus of archaeological data.

In response to this problem, attempts have recently been made to store and, in some instances, consolidate all archaeological datasets into single databases (e.g., BIAD [2]; Big Exchange [3]; EAMENA [4]; Pandora [5]). The fragmentary nature of the archaeological record means there are few opportunities for the truly 'big data' analyses that are more routine in other disciplines. These databases, alongside the body of published material, provide a rare chance to conduct such analyses. The use of artificial intelligence and big data models in scientific and commercial applications surged in popularity in recent years. Advances in deep learning, for instance, have enabled users to solve complex image processing problems. These new approaches are often argued to be the next frontier [6] in archaeology.

One of the richest sources of legacy data stored in published sources are illustrations and photographs of artefacts and archaeological features. These images preserve size and shape data, which can be used to make technological, functional, chronological, and cultural interpretations. In this paper, we apply AI and image processing complemented by a workflow for automatic recording of archaeological catalogues, demonstrating its utility with the analysis of graves and arrowheads. This AI-assisted and automated method of outline extraction makes marked improvements over previous image processing methods, which necessitated time-consuming pre-processing steps that constrained sample sizes, impeding the application of big data methods. While we test this workflow on burials and arrowheads, this workflow could be equally applied to the vast number of material remains for which outline size and shape holds techno-cultural significance.

## 1.2 Challenge: Consistent assemblages of archaeological data

Previous attempts to make sense of new and previously published photographs and illustrations have analysed manually extracted outlines from a wide range of artefact and ecofact types. For instance, quantifying outline shape from photographs has proven useful in the analysis of handaxes [7–14], points [15–19], flakes [20–26], and even botanical remains [27–31]. Meanwhile, illustrations have been used to compute outline shapes for items like lithic points [32,33], metal artefacts [34–36], and buildings [37]. Almost all of these studies require multi-stage manual pre-processing steps to re-format, scale, and clean the photographs or illustrations, often involving several image processing programs. Many of these studies also require the manual placement of landmarks for geometric morphometrics, which is time-consuming and prone to large inter-observer errors [38,39]. Automatic methods of outline extraction have the potential to make analyses of outline morphology considerably more efficient and objective, with larger sample sizes.

Incorporating legacy images, with various scales, resolutions, and orientations, that also routinely require cleaning of 'background noise', makes the collection of truly large sample sizes infeasible with existing methods. We introduce *AutArch*, an AI-assisted method, that automates many of these aspects of outline extraction to, for the first time, enable the efficient and objective processing of images from legacy data sources with an eye towards truly big data analyses.

Here, we outline the challenges that this AI-assisted workflow is aimed at addressing. When aiming to collate such assemblages, existing records cannot be integrated unchanged. Records with comparable quality, resolution, and consistently applied recording standards are required, which is rarely the case in archaeology. *Manually* standardising published archaeological illustrations would be a prohibitively time-consuming, repetitive and error-prone task.

A parallel problem is encountered when attempting to extract text-based information from the overwhelming corpus of published literature. Natural language processing (NLP) has served archaeologists for more than a decade [40–42] and is poised to change how we extract information from the ever-growing number of publications [43,44]. For instance, it has proven most useful in extracting information like chronologies, cultural taxonomies, and artefact types from grey literature field reports [42,45,46]. In our research reported here, we also aim at extracting data from legacy sources, but from images rather than text. Ultimately, text (NLP) and images (image processing) approaches could be combined, as we discuss as future work.

The aim of integrating images from thousands of archaeological PDFs is further hampered by difficulties in reliably and objectively finding relevant data. Screening thousands of pages of text and images to locate relevant content is a time-consuming and repetitive task; data can be scattered across many parts of the document; papers can be written in different languages and alphabets; various conventions for describing attributes such as the size and orientation of graves are inconsistently applied. Thus, the mere collation of data from published tables can result in inconsistent datasets. NLP has the potential to improve the integration of literature and datasets written in different languages [47]. But for images, using Computer-Assisted Design (CAD) or drawing software products to manually re-measure objects based on published drawings is tedious and prone to error. Therefore, large-scale assembly of uniform and commensurate data from a variety of sources is rarely possible in practice.

The lack of unified drawing standards in archaeology especially complicates data comparison across publications. Measuring the width, length and depth of a three-dimensional object, such as a burial pit, for instance, might be expected to yield consistent results independent of author and publication; yet, there are multiple ways to describe objects in archaeology, particularly for irregular shapes. Archaeological site manuals [48] rarely provide guidelines on how to measure non-symmetrical objects such as graves. And while some authors show how the length was measured by drawing a line through the grave [e.g. 49,50], the width is rarely indicated in the same manner. Even if this information was transparently reported, harmonising measurement conventions remains a challenge. Reporting the orientation of skeletons in graves is similarly fraught [51]. The orientation of graves is rarely given in degrees, and when it is, different methods apply, often confusing the orientation of the skeleton with the orientation of the grave cut.

Beyond graves, synthesising data from artefact analyses are hampered by a similar set of problems. There is little standardisation among published tables of artefact classifications or measurements.

Differences in recording standards and nomenclature become particularly pronounced when comparing different time periods. The measurements themselves are not always standardised either. For instance, there are even several ways of measuring something as simple as flake length, each resulting in different values [21]. And for any artefact, deviations in the angle at which the measurement is taken can result in considerably different length values [52]. Problematic inter-observer error rates in lithic, ceramic, and faunal analyses have long been well documented for manual measurements of linear dimensions [53,54], edge angles [55], and qualitative classifications [56]. Recent attempts at making comparative analyses more consistent [57,58] will help make future syntheses more robust, but legacy datasets remain fettered by mostly unknown amounts of inter-observer error.

In sum, there is no guarantee that reported measurements of objects can be compared across different publications. Whether measuring artefacts or an archaeological feature, like a grave, a uniform analysis of published images has the potential to remove many of the problems that otherwise confound syntheses of legacy data.

## 1.3 Contribution: An AI-assisted workflow for data collection

In this paper, we present *AutArch*, an AI-assisted workflow (and accompanying software) that allows fast, accurate and uniform identification and analysis of objects in unsorted PDF documents.

AI-assisted approaches are proliferating in many sub-disciplines of archaeology. In geoarchaeology for instance, machine learning has been used to more accurately source and classify samples of soils, minerals, and tephras [59–63], improve precision in temperature estimations of heat-treated lithic items [64], as well as to interpolate geochemical properties between samples to improve resolution of large-scale geological mapping [65]. These methods have also helped identify anthropogenic structures from aerial surveying [66], like mounds [67–75], structures [76–78], desert kites [79], irrigation systems [80], and combustion features [81,82]. The identification of anthropogenic cut-marks on bones has also been improved with machine learning [83–88]. As has the identification, classification, and interpretation of images like rock art [89–92], wall paintings [93], decorations [94,95], mosaics [96], and scripts [97–99]. Finally, machine learning has been instrumental in the analysis of the typology, style, and regionalization of archaeological artefacts like lithic points [100,101], ceramics [102–108], ivory carvings [109], glass vessels and coins [110].

These previous archaeological approaches leveraging AI have often addressed narrowly focused research questions aimed at specific features or artefacts. As of yet, the potential for AI-assisted workflows to be applied more generally and on larger sample sizes has not been fully explored. Despite archaeology's traditional emphasis on typological and geometric morphometric approaches [111], published resources such as archaeological drawings and photographs in illustrated catalogues have remained underutilised [112]. Yet, Hoggard et al. [113] demonstrated the reliability of employing legacy artefact illustrations from multiple sources in shape analyses. There is thus much potential for *AutArch* to provide a general solution that increases the usefulness of published resources by making their content immediately available to archaeologists as data in a well-defined format.

## 1.4 Testing Ground: Third millennium BCE Europe

Due to the 'messy' [114], but to some extent repetitive nature of the archaeological record of third millennium BCE Europe, this period provides an interesting testing ground for innovative AI-assisted approaches to data collection. Tens of thousands of graves and grave inventories have been ascribed to this period. Historically, archaeological assemblages have been grouped into archaeological 'cultures' and 'groups', based on different classes of attributes, making comparison difficult. Southeast European 'Yamnaya' assemblages, for instance, are defined after a specific type of burial, Russian 'Yama', i.e. 'pit'-graves [115,116]; North-Central European 'Corded Ware' assemblages are characterised by a specific ceramic decoration [117]; Western European 'Bell Beakers' are named after the shape of the ceramic vessels found in graves [118,119].

One of the most common criteria used to distinguish 'Corded Ware' graves from 'Bell Beaker' ones is the orientation of the skeletons in graves, usually East-West versus North-South in Continental Europe [51,120,121]. However, graves are complex, irregular objects that are rarely scaled and oriented in publications. Graves are destroyed during excavations and cannot be digitized retroactively, making the use of legacy resources such as catalogues essential. The findings are frequently reported in the form of technical drawings and photographs, accompanied by descriptions written following a variety of professional standards in different languages. Unlike artefacts, graves are typically assigned a name or burial number, enabling querying of image and text in publications.

The funerary record from the third millennium BCE appears to be a strong candidate for the challenge of developing an automated workflow that requires only minimal input from researchers to digitise large datasets recording rich, numerical annotation. On this testing ground, we evaluate the research results along several dimensions; object detection is evaluated in terms of standard metrics; the workflow is evaluated by means of metrics quantifying automated versus manual steps for (a subset of) our dataset; correctness and productivity (usability) of the workflow, as supported by the *AutArch* software, is evaluated by a user study, which also covers comparison with standard practices (such as *Inkscape* for measuring graves).

## 1.5 Online Resources

For reproducibility, the paper is complemented with online resources (**Supplementary Material S1**). We provide an easy-to-use copy of AutArch with all publicly available assets under: https://doi.org/10.5281/zenodo.14999893. This version includes the database, but not the supporting PDF images, which are protected by copyright. Please contact us if you have any questions regarding the operation of the software. The complete AutArch source code is available in the following GitHub repository: https://github.com/kevin-klein/AutArch.

# 2 Materials and Methods

## 2.1 Object Detection

### 2.1.1 General Object Detection Setup

In image processing, two types of neural networks with different purposes are commonly used: object detection, and classification (**Supplementary Material SI 2**). Classification assigns an object, such as an image of a skeleton, to a predefined number of labelled classes, such as 'supine' or 'flexed on left side', to describe the pose of the skeleton in the grave. Object detection networks, on the other hand, output a list of bounding boxes with a label for each of them. Each bounding box corresponds to a specific object detected in the image.

Models need to be trained to accurately perform the desired task. For supervised training of a model, one needs a training dataset, a loss function and an optimiser. In our case, the training dataset consists of images of objects that were manually labelled and that the algorithm learns to recapitulate. We use two different types of models in our application. We decided to use RetinaNet, because it is currently the best performing object detection network available in PyTorch [122,123]. Other implementations were not considered due to potential future maintenance issues. The exact choice of object detection network is not integral to the AutArch workflow. We also provide other trained model data as alternatives to RetinaNet in the GitHub repository. Other implementations might be preferred for practical reasons, such as deployment targets. The second network is Resnet (residual neural network) [124], which consists mostly of stacked convolutional layers, a common architecture for image classification networks, and represents a significant improvement over previous networks mostly as it introduced shortcut connections.

### 2.1.2 Object Detection for Graves in Legacy Sources

Documents in PDF format were converted into images and then were fed into RetinaNet trained with a set of archaeological drawings (**Supplementary Material SI 2**). This results in a list of bounding boxes with labels for each page. These labels include “grave”, “skeleton”, “scale” and “north arrow”, among others. These bounding boxes encapsulate the relevant object as shown on the page (**figure 1**). On their own, the objects detected in individual bounding boxes are of little use to archaeologists. Different classes of objects have to be combined to represent measurable and interpretable entities. This means that graves have to be connected with their respective scales, north arrows, cross sections, artefacts and possible skeletons contained within the grave.

While the object detection network is trained to detect both vectorised images and photographs of graves and artefacts, important aspects of *AutArch*, such as outline extraction, are not currently optimised to work with warped objects in photographs. Therefore we recommend using *AutArch* on nadir images taken at a 90-degree angle to the ground with lens distortion correction. Heavily edited photographs of artefacts, produced in a controlled environment with good lighting, high resolution, high contrast, such as those described in section 4.1.3, should yield similar good results.

To create fully annotated graves, the algorithm first combines all graves on a page. The nearest north arrow, scale and cross-section are assigned to each grave, using the Euclidean distance measured from the centre of the bounding boxes. These objects can be assigned to multiple graves at the same time. All skeletons and artefacts within the bounding box of the grave are then assigned to it.

Scales have to be further processed to parse the information they encode. For this, we devised a method using the contours of the bounding box and OCR (optical character recognition) with Tesseract [125] (**Supplementary Material SI 2**). Tesseract was used for this purpose, because it was the only open source OCR solution that had a ready-to-use ruby binding [126] with the rails backend. Commercial alternatives were not considered at this stage, because of the scope of the AutArch project and its ambition to make any software available under a non-restrictive licence. The contour analyser calculates the length of the scale, while we use OCR to determine the factor for conversion from pixels into real-world distance. The text associated with the scale, usually a number with a length unit, is extracted, and then the software tries to determine the unit of measurement, such as centimetre or metre.

The north arrows are analysed using a Resnet-152 [124] residual neural network that retrieves their angle. This network has been trained in an unsupervised manner as opposed to other neural networks discussed here. For further details, see **Supplementary Material SI 2**. Resnet-152 was preferred, because of its popularity ensuring widespread use [127]. It is one of the neural networks officially supported by PyTorch [128] and it uses an updated architecture to improve its performance [129].

The contours of objects are extracted in the same manner as scales. Then the largest contour is selected using the length of its arc. This outline is used to calculate the area, width, length and rotation of the burial. The area is retrieved using the OpenCV contourArea function. Width and length are created by first obtaining the bounding rectangle using minAreaRect [130]. By definition, the width is the shorter side of the rectangle and length is the longer side. The angle is also derived from the bounding rectangle. The poses of skeletons are classified with Resnet-152.

Several challenges arose when using the aforementioned method on a wide variety of publications. Occasionally, objects are incorrectly assigned to a grave due to proximity. Often this is because multiple scales are present on the page and the nearest scale belongs to an artefact instead of a grave. False negatives can also occur, leading other objects to be incorrectly grouped with a grave, such as north arrows belonging to a different drawing on the same page. In such cases, a domain expert must make manual changes as needed (see **4.3**). Manually redrawing a contour reduces the speed of the overall process by a few seconds. The process has been designed to be as familiar as possible to users using hermite splines, as used in common vector drawing software.

A list of all graves in a publication is automatically generated once the process is complete. Each entry must subsequently be manually validated.

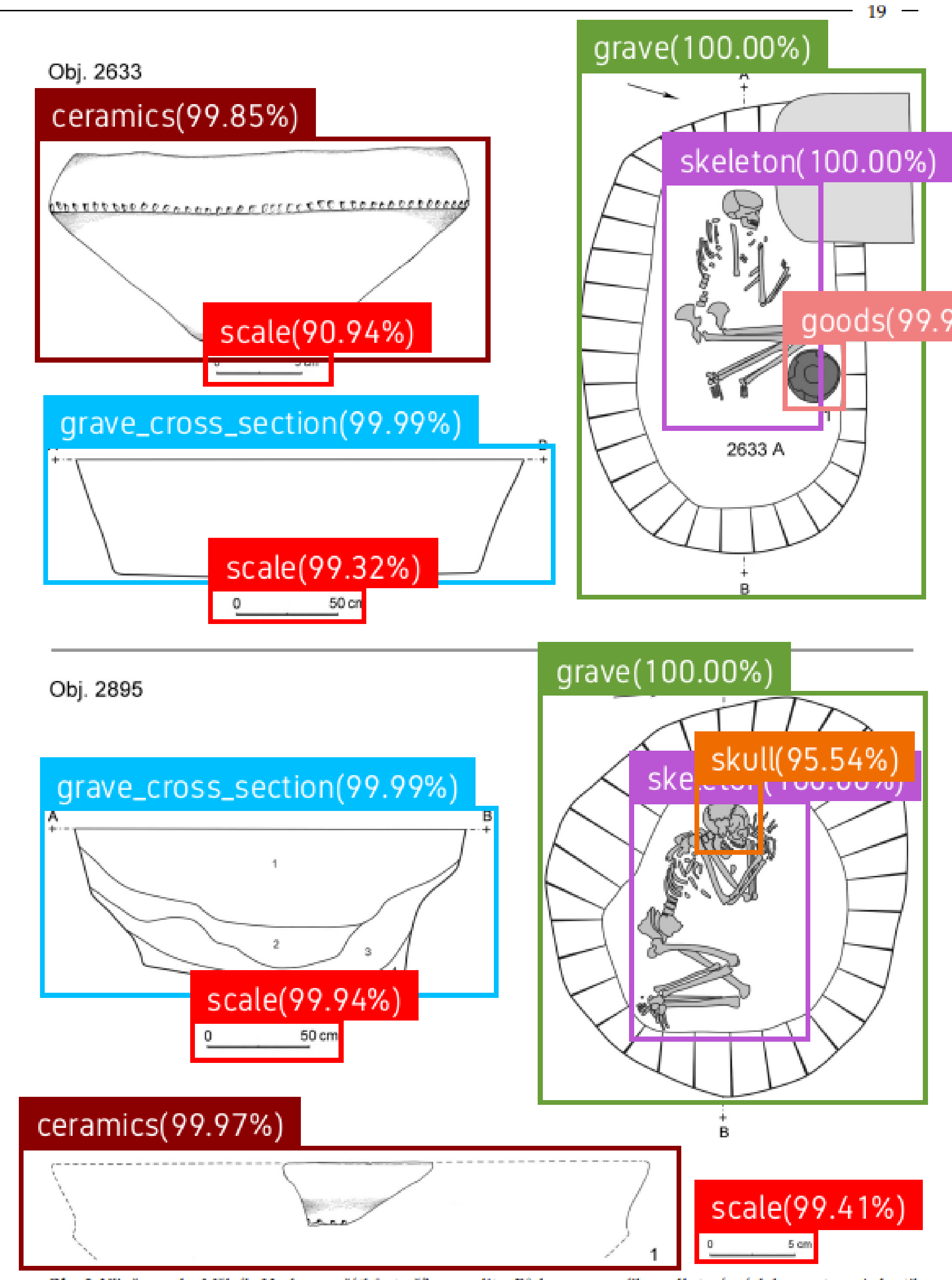

***Figure 1***. *Example of the object detection result for one page of catalogue (here: Corded Ware site of Vliněves* [1]*, Czech Republic). Results for North arrows were excluded for better visibility.*

## 2.2 Datasets

The primary dataset is the one for detecting graves and objects therein. Two derived datasets are also considered – they deal with skeleton pose and arrow-type classification for skeletons and arrows detected via the primary dataset.

### 2.2.1 Object Detection Training Dataset

As a primary dataset, we used catalogues of 3rd millennium BCE graves, which we manually annotated for object detection using *labelImg* [131] (**Supplementary Material SI 3**). In total, 391 pages were annotated (**table 1**). The quality and detail of the annotated material range from high-resolution digital catalogues, such as those published for the sites of Vliněves in the Czech Republic [1,132], to lower-quality book scans and even hand-drawn notes from Southeast European and Russian excavations. The corresponding archaeological sites are shown in **figure 2**.

Each page of these documents was converted to an image and randomly sampled. At first, we annotated all pages with drawings from four publications for the initial proof of concept [1,132–134]. For the extended dataset, a random sample was used and a manual selection was done to maximise the number of pages containing relevant drawings.

| **Class label** | **Object count** | **Scanned hand-drawn images** | **Scanned digital images** | **Digital illustrations** |
|---|---|---|---|---|
| grave | 376 | 72 | 82 | 222 |
| text | 850 | 174 | 116 | 560 |
| skeleton photo | 52 | 6 | 0 | 46 |
| ceramics | 615 | 132 | 128 | 355 |
| artefact | 535 | 342 | 99 | 94 |
| grave photo | 122 | 4 | 2 | 116 |
| map | 59 | 9 | 11 | 39 |
| scale | 905 | 236 | 155 | 514 |
| arrow | 532 | 160 | 118 | 254 |
| skeleton | 404 | 146 | 91 | 167 |
| grave artefact | 207 | 8 | 26 | 173 |
| grave cross section | 240 | 44 | 19 | 177 |
| stone tool | 346 | 112 | 180 | 54 |
| shaft axe | 121 | 2 | 111 | 8 |

***Table 1***. *Number of annotated objects in the primary dataset per class and image quality (see supplementary material for list).*

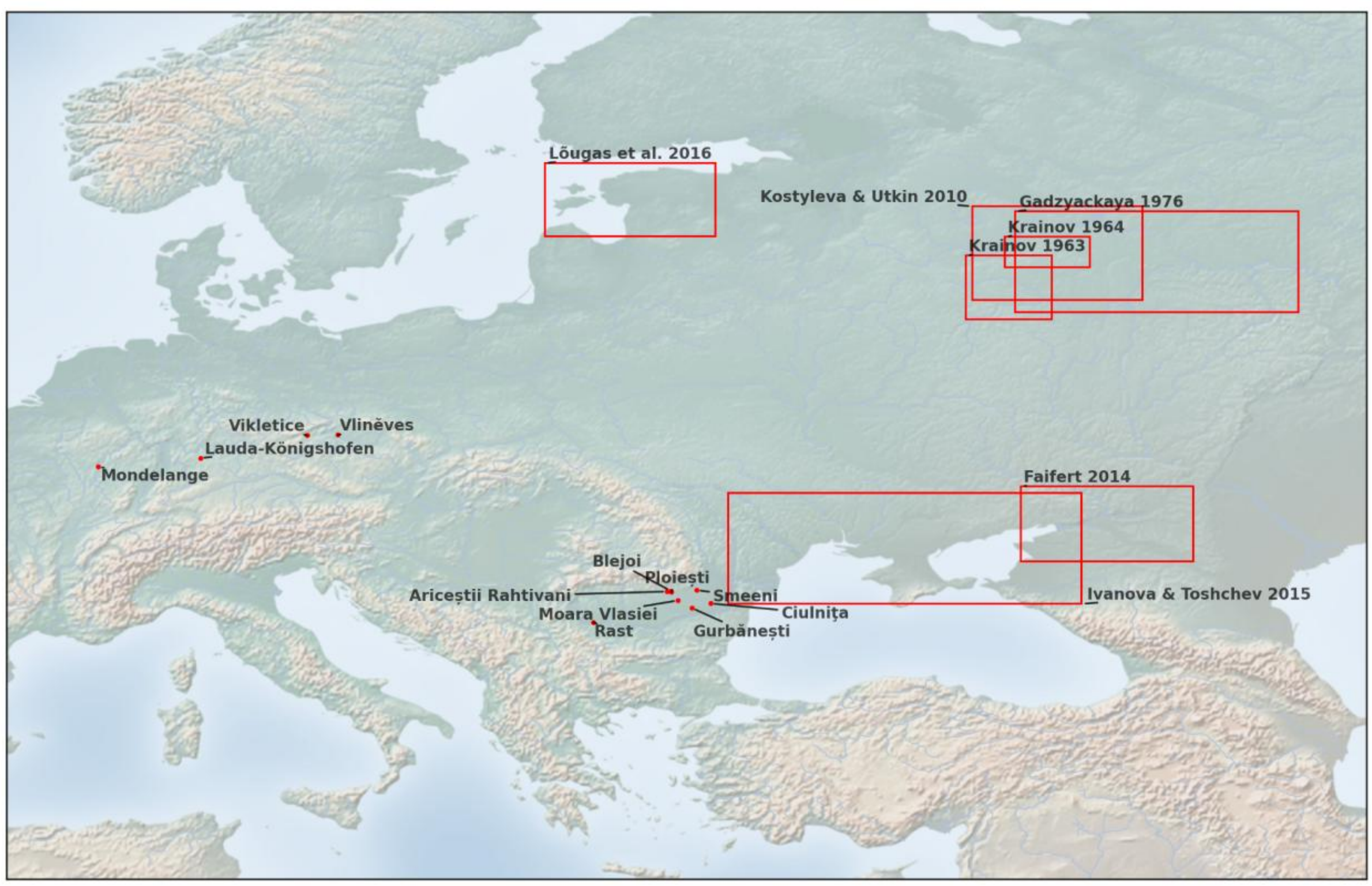

***Figure 2****. Map of sites used in training (drawn with Matplotlib Basemap Toolkit 1.3.8 and Python 3.11; background is a display shaded relief image from [http://www.shadedrelief.com](http://www.shadedrelief.com)).*

### 2.3 Other Datasets

The datasets used for the training of the skeleton pose and arrow classification are further documented in **Supplementary Material S2**.

# 3 The AI-assisted Workflow

## 3.1 Workflow Steps

The full semi-automated workflow for annotation consists of eight distinct steps – with appropriate support by the graphical user interface of the *AutArch* software:

*Step 1. Basic grave information.* The ID assigned to the burial by the authors in the source publication is recorded. In case multiple images of the same grave are shown, the software will prevent duplicates in the results using this ID. In this step, the expert also has the option to discard drawings incorrectly classified as a grave.

*Step 2. Site.* Optionally a grave can be assigned to an archaeological site. Sites are identified by their general location and name.

*Step 3. Tags.* A grave can be assigned arbitrary tags for filtering purposes.

*Step 4. Correcting bounding boxes.* The user can manually add, remove or change the bounding box assigned to a specific grave. Potential tasks include selecting a different scale on the page, resizing bounding boxes because they do not fully encapsulate an object or marking north arrows that were initially missed by object detection. During this step, a manual arrow has to be drawn for every skeleton following the spine and pointing towards the skull, which is necessary to determine the orientation of the skeleton in the grave. Several automated steps are then performed. The contours are calculated using the new bounding boxes and the resulting changes in measurements are saved. The orientation of the north arrow and the deposition type of the skeleton are updated using their respective neural network. The analysis of the scale is performed again.

*Step 5. Contour validation.* All detected outlines in relation to one particular grave are highlighted, allowing the user, if any issue arises, to return to the previous step and fit, for instance, a manual bounding box around the grave or cross-section to indicate the width, length or depth.

*Step 6. Scale information.* The next step is to validate the scale by checking the text indicating the real-world length of the scale. Once this step is completed, all measurements are updated with the new scale information. In case no individual scale is provided and the publication uses a fixed scale, e.g. all drawings are 1:20, a different screen is shown. In this screen, the actual height of the page (in cm) has to be entered manually, together with the ratio of the drawing. This way, all measurements can be calculated in the absence of a scale and the results are fully compatible with scaled publications.

*Step 7. North arrow orientation.* The angle of the north arrow can be adjusted manually based on a preview. In case an arrow is missing in the drawing, this screen will be skipped and size measurements and contours will still be collected without the orientation.

*Step 8. Skeleton pose.* Finally, the pose of all skeletons has to be validated, which (for now) consists of "unknown", "flexed on the side" or "supine". As described above, a neural network will set the initial body position, but it can be adjusted manually. Further positions could easily be added in the future. "Unknown" is used in cases where skeletal remains are visible, but no position can be identified.

## 3.2 Manual Validation and Adjustment

The workflow is supported by a graphical user interface (as part of the *AutArch* software) that allows manual correction of the annotations obtained through the automated process described above. Although our automatic approach is accurate for most publications, validation and potential manual correction by a domain expert are essential to maximise the reliability of the archaeological annotations. The graphical interface is a web application that is usable on a wide variety of devices including mobile devices. While the initial focus was on drawings of graves, artefacts, such as arrowheads, can be analysed with minor modifications, as demonstrated in 4.1.3.

Our two-part workflow uses an automated process and a manual process. In **figure 3**, we illustrate manual adjustment, as supported by the *AutArch* graphical user interface.

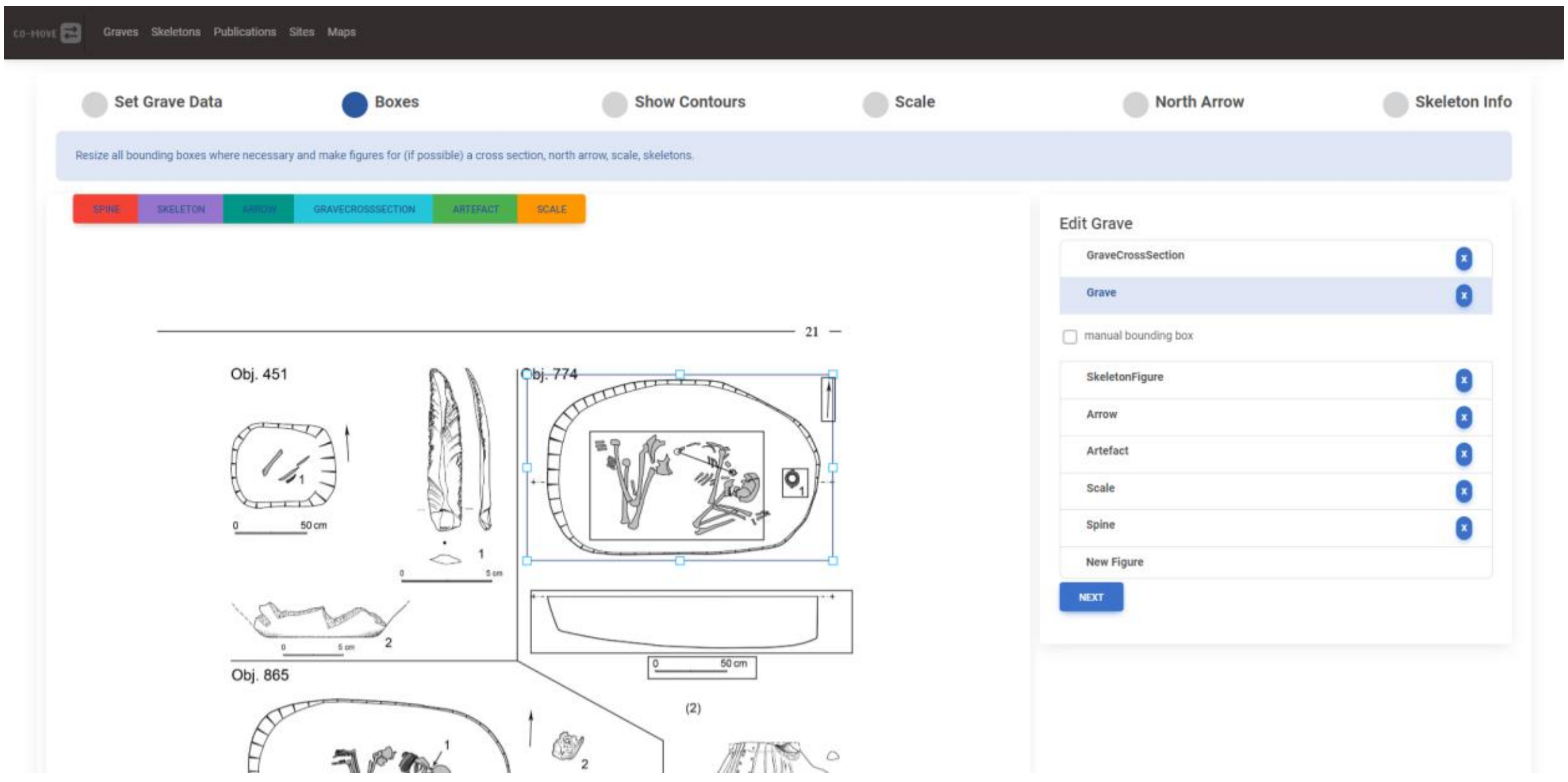

***Figure 3.*** *Screenshot of the manual validation process; note the resizing of the box around the grave.*

# 4 Results

In this section, we discuss results obtained through *AutArch* and their archaeological implications, the evaluation of object detection, workflow, and the achievable user experience. The data related to this evaluation is published as part of the *online resources* (**Supplementary Material S1**).

## 4.1 Data Analytics and Visualisation

Besides the step-by-step workflow, the *AutArch* software also supports data analytics and visualisation of summary data for multiple graves or publications in terms of the orientation of skeletons or the geometry of outlines. Our method of aggregating archaeological data is tailored towards supporting the creation of uniform assemblages. Publications are visualised as a whole with a variety of charts. Several publications can be compared in the software. In this case, the data is usually presented in the same chart and is assigned a specific colour that is consistent through different visualisations. All data points are linked to their respective burials and can be navigated.

### 4.1.1 Orientation of Skeletons

*AutArch* performs descriptive statistical analysis for data visualisation purposes. The workflow can automatically generate scientific maps showing the distribution of attributes, such as the orientation of skeletons in graves using radial charts (**figure 4**). For instance, when comparing burials attributed to the 'Corded Ware' [1] culture at Vliněves with those attributed to the 'Bell Beaker' [132], we observe differences in the orientation of the skeletons (**figure 4**).

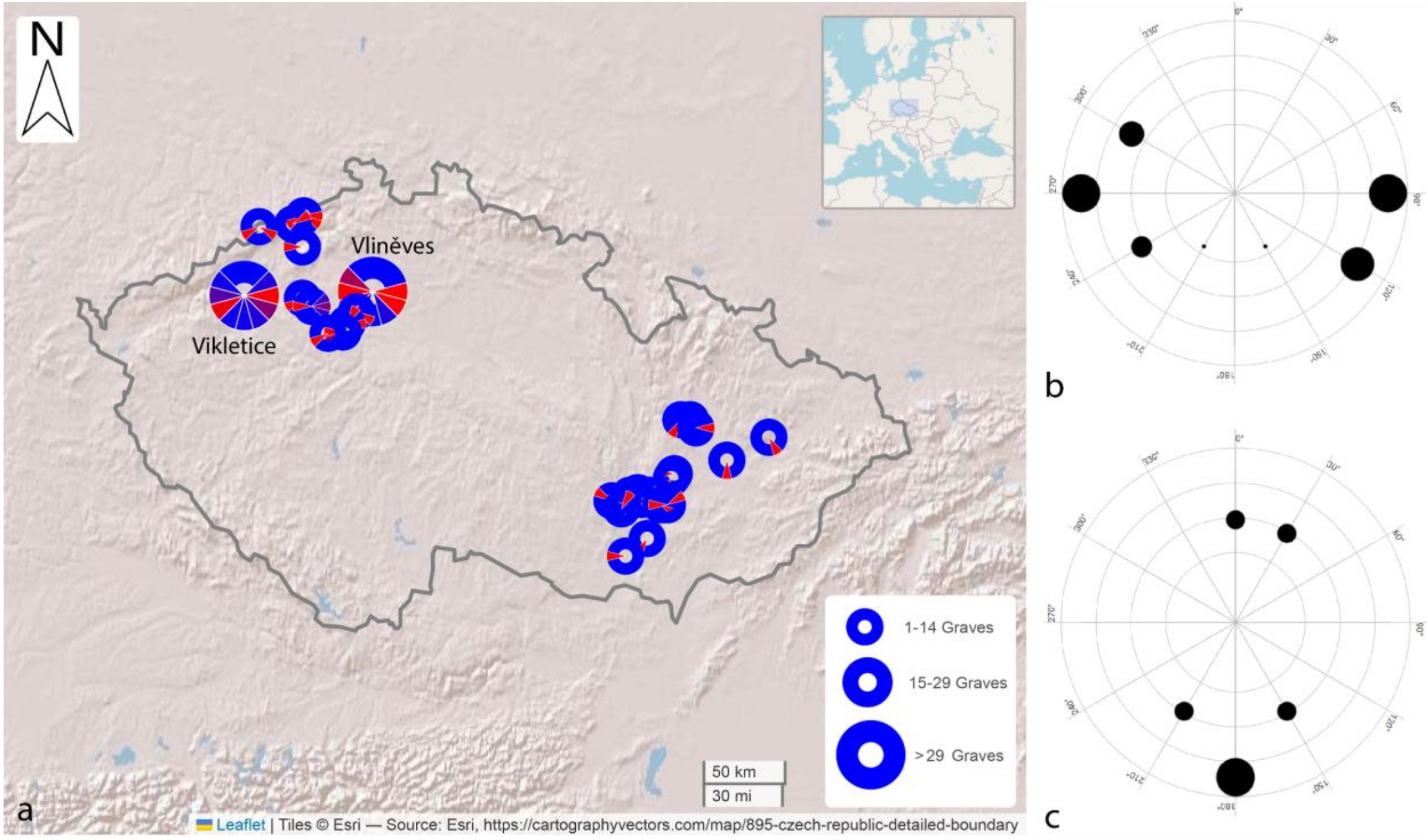

***Figure 4***. *An example of numerical attribute that can be automatically retrieved using AutArch: the orientation of graves with skeletons from (a) 'Corded Ware' sites of the Czech Republic (n = 147); (b) 'Corded Ware' levels at Vliněves [1] (n = 39); and (c) 'Bell Beaker' levels at Vliněves [132] ( n = 6).*

### 4.1.2 Geometry of Grave Outlines

Outlines of graves can be stacked, scaled and oriented using *AutArch* to visualise their shape (**figure 5**). Drawings that have no automatically detectable outline and use the manual bounding box instead are excluded here. We compared 109 grave outlines automatically extracted, scaled, and oriented from Vliněves using *AutArch*, including 78 Corded Ware [1] and 31 Bell Beaker graves [132]. Elliptical Fourier Analysis (EFA) was conducted with these outlines in PAST [135], with 20 harmonics projected onto two dimensions using a Principal Components Analysis (PCA). The orientation of Corded Ware and Bell Beaker graves are noticeably different in **figure 5c**, as already discussed above. However, an EFA, invariant to orientation, shows that there is no discernible difference in outline shape between Corded Ware and Bell Beaker graves at these sites (**figure 5a**), underscoring their close cultural affinity [136].

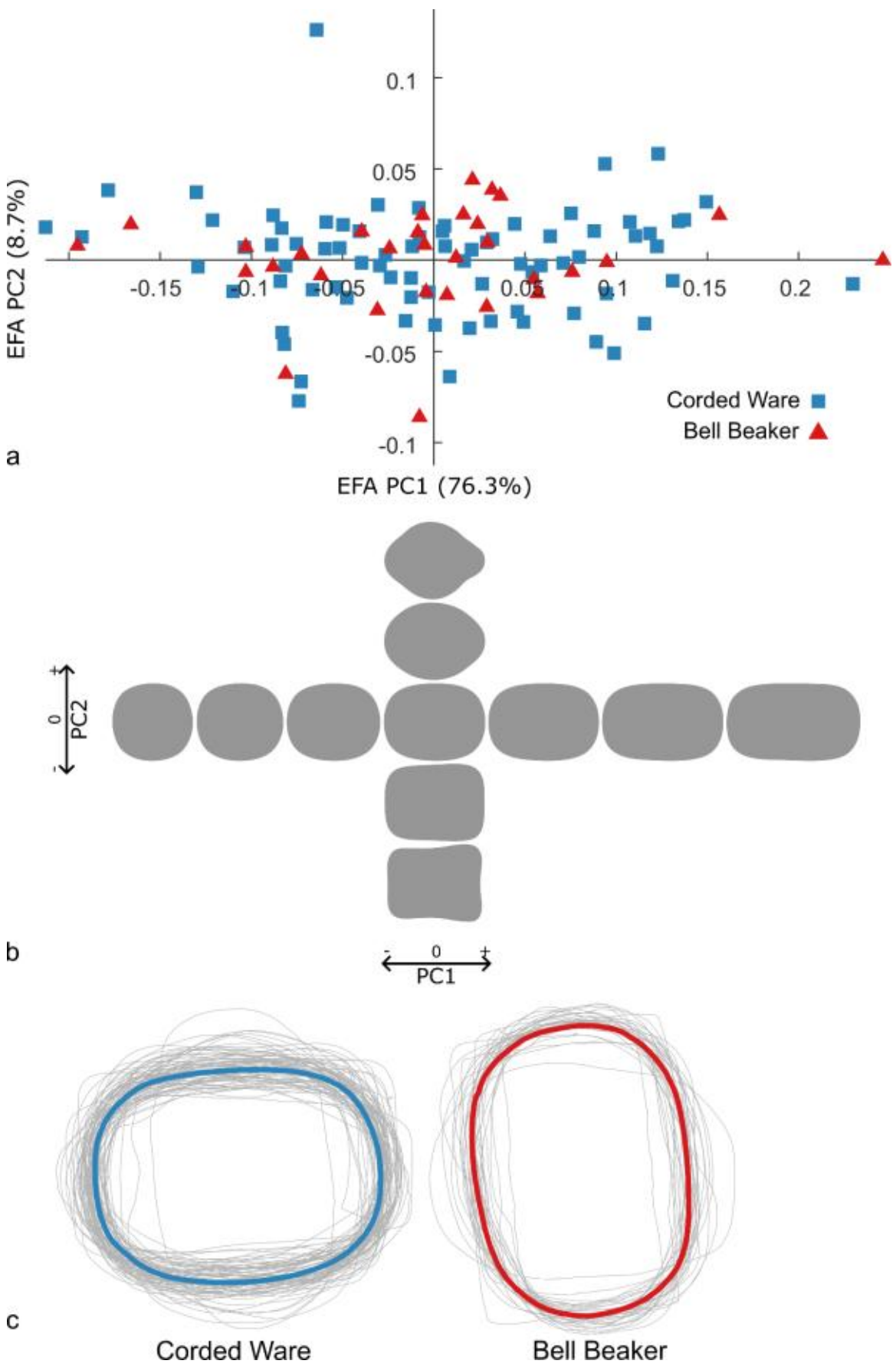

***Figure 5.*** *Elliptical Fourier analyses of a sample of grave cuts extracted using AutArch. (a) PCA of the EFA coefficients of Corded Ware and Bell Beaker grave cuts; (b) Reconstructed approximation of grave shapes using elliptical Fourier descriptors, plotted along the same axes as in a, in increments of 0.05; (c) Closed contours of Corded Ware and Bell Beaker graves in grey, with mean shapes overlain in colour.*

### 4.1.3 Geometry of Artefact Outlines

To demonstrate the potential generalisability of *AutArch* beyond the study of graves, we provide a test in the Supplementary Materials on the program's ability to extract artefact outlines. There have already been a number of recent attempts at automating parts of the artefact outline extraction process. Some, like Lithics3D[137], Artifact3-D[138,139], and CEAM[140] require 3D scans as input, which means they are, for now, constrained by the time and cost of 3D scanning, precluding big data approaches. Other programs, like outlineR[18], PyLithics[141], and PyREnArA[142], extract outlines from 2D images. The vast quantity of published illustrations available makes these methods more scalable and prone to big data methods. PyREnArA and outlineR focus on automating steps of shape analysis, but involve time-consuming pre-processing steps. Meanwhile, PyLithics automates parts of the outline extraction process for stone tools, but involves artefacts drawn following specific illustration conventions. Although this allows a great deal of information to be extracted, including scar sizes and directions, our goal here is to extract larger amounts of commensurate data from more diverse sources. For now, we mainly extract outlines, but can do so for all illustrated archaeological material, not just lithics and not just artefacts. Given the diversity of illustration conventions, even within a class of artefacts, there is a need for methods that can process images irrespective of archaeological material, a task that AI-assisted workflows are well suited to address. While other workflows offer ways of automating some analysis steps, they first require manual processing steps, many of which are automated by AutArch, which also automatically identifies the visualized objects in the first place. Yet, as these different software each tackle different portions of the extraction and analysis workflow, they can be used together to address specific samples (e.g. PyLithics) or different steps of analysis (e.g. PyREnArA or OutlineR). The choice to use these different software comes down to the sample one wants to analyse. AutArch is suited for those seeking to automatically identify, extract, scale, orient, and measure visualizations of any archaeological material.

We tested the potential of *AutArch*'s outline extraction process on a large sample of arrowheads presented by Nicolas[143] (SI 5). We replicate the findings of Nicolas[143], as well as those subsequently quantified by Matzig et al.[18], who both found that the arrowheads of the Neolithic and Bronze Age of Northwest Europe show strong spatiotemporal patterning (**supplementary figure 3**). The extracted arrowhead outlines derive from both photographs and illustrations, of varying resolutions, backgrounds, and illustration conventions, demonstrating the versatility of the object detection and outline extraction method.

### 4.1.4 Implications for the Archaeology of Third Millennium BCE Europe

The reliability of expert classifications, such as 'Corded Ware' and 'Bell Beaker' can be validated using grave and artefact shape variations. The focus on shape obviously limits the scope of this study to one aspect of the objects. However, it nonetheless provides wholly neutral data, stripped of all context and interpretation. As demonstrated in **Figure 5**, grave outlines are relatively simple shapes and do not show much variation when projected on PCA axes, supporting notions of a 'Single Grave Burial Ritual' [136]. For artefact analysis, we have for now applied *AutArch* on the arrowheads of the Neolithic and Bronze Age (**Supplementary Figure 3**). Artefacts like arrowheads and backed artefacts are among those most amenable to these sorts of analyses, as their shape in plan-view holds technological [144,145], functional [7,146], and cultural [18,19,147,148] significance. Yet, *AutArch* could equally be applied to any number of artefact types. Further iterations of *AutArch* will help move away from our current interpretative dependency on expert classifications of cultural labels by archaeologists and have the potential to create new, empirically-meaningful and statistically robust, 'prehistoric' taxonomies [149] for third millennium BCE Europe and beyond through machine learning.

For example, our study confirms the existence of binary oppositions in body orientation within 'Corded Ware' and 'Bell Beaker' graves of the Czech Republic, which conform as predicted in 1.4 to overall East-West versus North-South orientation patterns (**Figure 6**). Exact angles, derived directly from the drawings, are provided for each of these categories, improving on previous classifications based on expert knowledge or data extracted from tables. The current picture overlooks significant individual and regional differences, such as those between Bohemia and Moravia (**Figure 4**), which can now be accurately captured using *AutArch*. The prospect for archaeology is to move from expert-made maps of the burial record in 3$^{rd}$ millennium BC Europe, such as those produced by Häusler [150] and more recently Furholt [151] to visually convey information about prehistoric cultures, to scientific ones based on large-scale data, collected in a consistent manner.

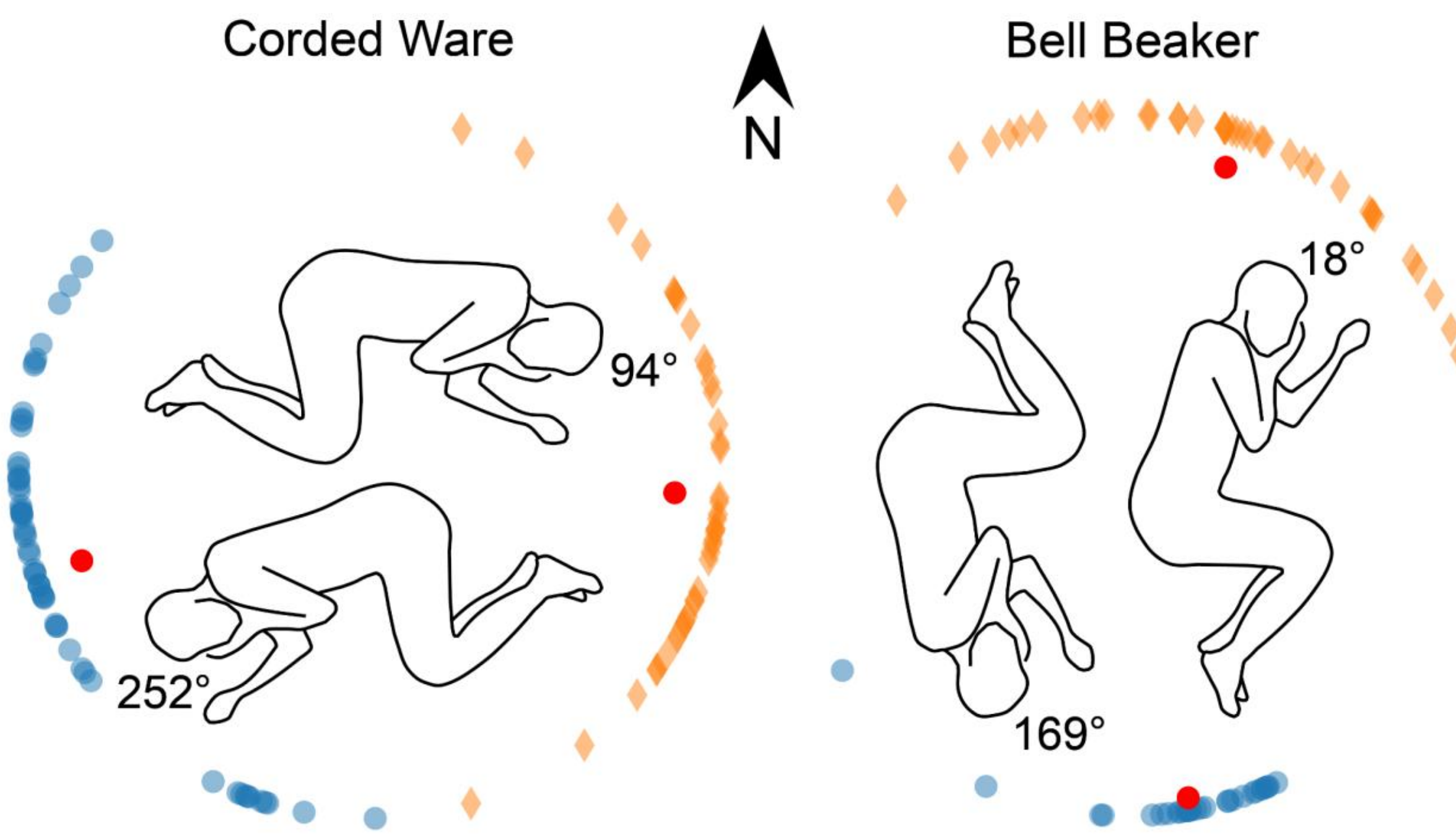

***Figure 6***. *Burial orientations based on 100 Corded Ware graves and 66 Bell Beaker graves with skeletons that have a discernible orientation from the Czech Republic analysed with AutArch. Binary differentiation assuming distinct burial orientations for females and males [51,114,120,152,153] was made with k-means using 2 centroids (red dots); silhouette score [154] for Bell Beaker: 0.78, Corded Ware: 0.68. The angle was calculated using the circular mean. Body figure drawings adapted from Bourgeois and Kroon 2017 [152].*

## 4.2 Evaluation of Object Detection by Metrics

To evaluate the software's ability to detect graves in archaeological publications, we performed a qualitative analysis on publications, which cover both related and different archaeological periods compared to our training dataset, including European Early Neolithic, Chalcolithic and Early Bronze Age graves.

As **table 2** illustrates, object detection performed perfectly for two of the four publications, whereas there were false positives and negatives (i.e., incorrectly detected or missed graves) affecting approximately 1/4 of the total number of graves for the two remaining publications. It is worth noting that the current training dataset of just 391 pages is rather small when compared to other commonly used datasets such as COCO 40. Additional training will further improve the object detection rate.

| Publication | true positive | false positive | false negative |
|---|---|---|---|
| Włodarczak 2018 [155] | 13 | 0 | 0 |
| Baron et al. 2019 [156] | 2 | 0 | 0 |
| Sachsse 2015 [157] | 53 | 2 | 12 |
| Neugebauer-Maresch and Lenneis | 84 | 22 | 4 |

| 2015 [158] | | | |
|---|---|---|---|

**Table 2.** Object detection results for graves in different publications.

## 4.3 Evaluation of Workflow by Metrics

As an estimation of the necessary manual work, we tracked all changes that needed to be performed to process Włodarczak 2018 [155] using *AutArch* (also mentioned in **table 2**). Only steps that could have been correctly performed by the automated step have been recorded. All graves were correctly detected in Włodarczak 2018 [155], so no manual marking of burials was necessary in a separate previous step.

As **table 3** illustrates, the automation of the different workflow steps occasionally fails, requiring manual assistance. For this sample, approximately three manual corrections are required per grave on average. Still, the guided workflow and the tool support for performing such manual adjustments further contribute to productivity and consistency.

| **Site** | **Grave** | **AS** | **AA** | **ST** | **AF** | **MB** | **DT** | **GB** | **AMS** | **AAD** | **Sum** |
|---|---|---|---|---|---|---|---|---|---|---|---|
| Prydnistryanske | III/3 | ● | ● | ● | ● | | | | | | **4** |
| Porohy | 3A/7 | ● | ● | | | ● | ● | | | | **4** |
| Kuzmin | 2/2 | ● | ● | | ● | | ● | ● | | | **5** |
| Ocniţa | 6/24 | | ● | ● | | | | | ● | | **3** |
| Ocniţa | 7/14 | | ● | ● | ● | | | ● | | | **4** |
| Tymkove | 5 | | ● | ● | ● | | ● | | ● | | **5** |
| Porohy | 2/5 | | | | | ● | | | | | **1** |
| Pidlisivka | 1B | | | | | ● | | | | | **1** |
| Severynivka | 1/5 | | | ● | | | ● | | | ● | **3** |
| Porohy | 3/2 | | | ● | ● | | ● | ● | ● | | **5** |
| Pysarivka | 6/2 | | | | ● | ● | ● | | | | **3** |
| Pysarivka | 8/2 | | | | ● | | | ● | | | **2** |
| Pysarivka | 5/1 | | | | ● | ● | | | | | **2** |
| | **Sum** | **3** | **6** | **6** | **8** | **5** | **6** | **4** | **3** | **1** | |

***Table 3.*** *Necessary manual corrections for Włodarczak 2018 [155]. Abbreviations: AS = adjusted scale, AA = added north arrow, ST = text entered for a scale, AF = north arrow flipped by 180°, MB = manual bounding box, DT = entered deposition type, GB = resized grave bounding box, AMS = added missing skeleton, AAD = arrow adjusted.*

## 4.4 Evaluation by User Study

To evaluate the correctness and potential productivity benefits of the software and the workflow, we invited 10 participants to perform small computing tasks for 90 minutes at the Anthropological Institute of Johannes Gutenberg University Mainz. This user study demonstrates that our AI-assisted workflow was significantly faster and less prone to errors than manual measurement using drawing software (**figure 7**). We expect this effect to become more pronounced the longer these tasks are performed, due to the onset of mental fatigue. **Supplementary Material S4** has complete details about the experiment.

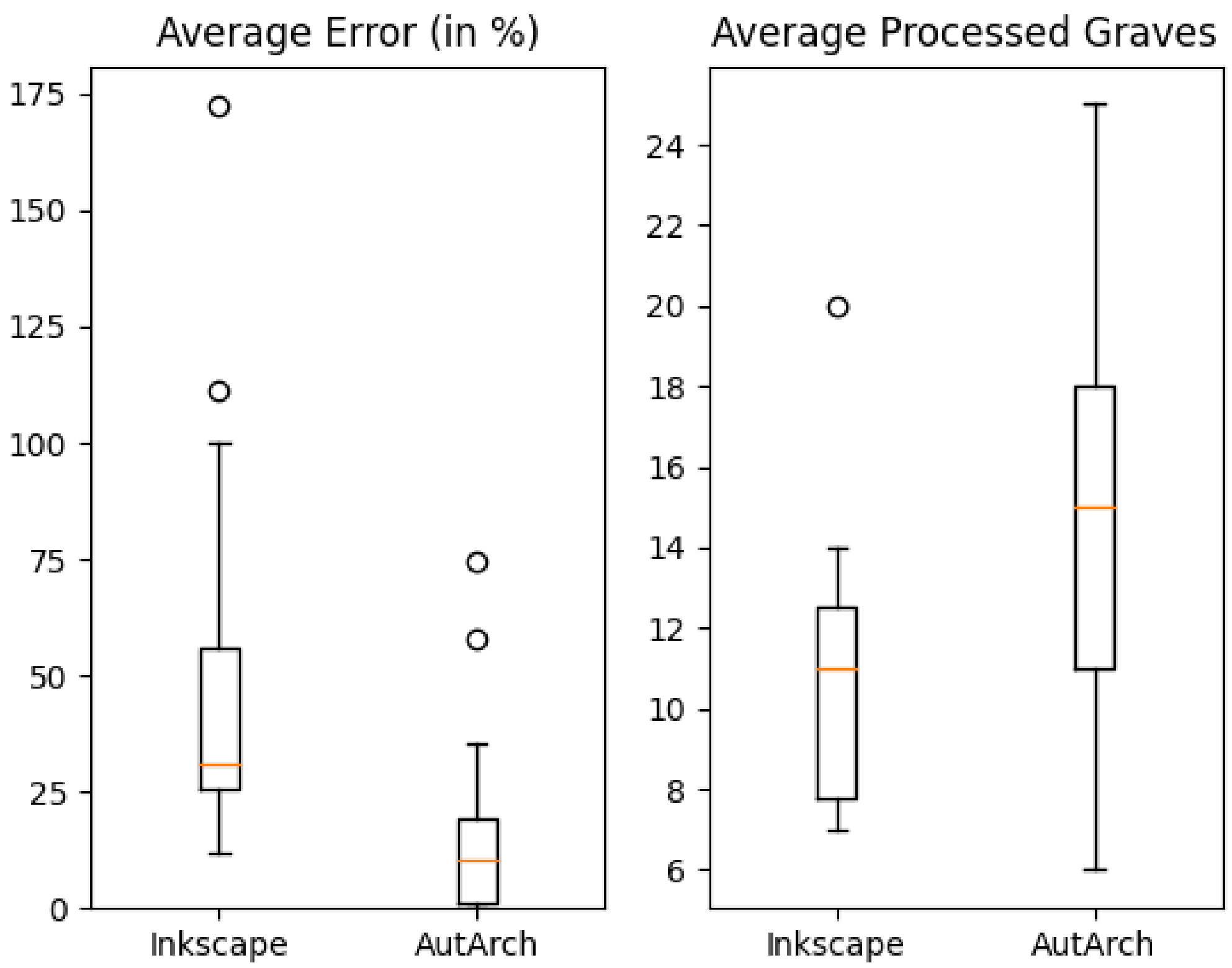

***Figure 7***. *Results of the user study. See* ***Supplementary Figures 1-2*** *for a breakdown of the results per user.*

# 5 Conclusions

## 5.1 Summary

We have presented a new AI-supported workflow to tackle the challenge of large-scale archaeological data collection. Our approach improves the speed and accuracy of manual methods via the use of AI-based object detection and image processing, as well as the use of a software-based workflow (*AutArch*). Also, our shape-related measurements of width, length and depth of burial pits are collected in a standardised manner, allowing the performance of meta-analyses across different publications. Our approach also increases the precision of measured angles to assess the orientation of skeletons and burial pits. Furthermore, the automated detection of contours can speed-up geometric analyses of a range of archaeological materials, demonstrated here with arrowheads.

The emerging database will allow for composite queries over all recorded attributes, such as burials within specific dimensions, with specific orientations or with a certain number of skeletons, as well as artefacts of certain sizes or shapes.

## 5.2 Future Work

To enable and encourage wider use, we plan to deploy AutArch on a server with web access and a service-level agreement (SLA) that is acceptable for research purposes. Since the input drawings are technical in nature, representing factual data rather than creative choices, they do not fall under copyright law. The EU directive 2019/790 of the European Parliament and of the Council of 17 April 2019 on copyright and related rights in the Digital Single Market and amending Directives 96/9/EC and 2001/29/EC provides a legal framework for data mining scientific publications. Other countries such as the UK provide comparable fair use copyright exceptions for non-commercial research [159]. The AI-assisted workflow will be shared with the community through an open science collaborative online platform, allowing continuous integration of material and cross-validation of studies relying on the integrated materials. Longer term, we plan to expand the method's coverage in terms of archeological objects and forms of processing, especially integrating NLP to combine information from images and text. Workshops and seminars devoted to complete case studies will help to reach critical mass and exploit the method for new research questions.

There is huge potential to improve data collection in archaeology for other objects beside graves and arrowheads, including but not limited to buildings, ecofacts, ceramics, metal tools and weapons. *AutArch* could be used as an excavation analysis tool in commercial archaeology. Even large amounts of unpublished hand drawings could be analysed and integrated into a database. A new standard for documenting archaeological features, like graves, might need to be created to ease integrations with software that can automatically digitise handwritten documentation.

Applying our method to other aspects of material culture, for instance, ceramic ware, would require a broader set of attributes. These would include, for instance, the fabric and surface treatment. Published text might contain further information, but its analysis will be a future task. Analysing the text of publications using established statistical NLP [40,41,43–45] or incorporating text into image classification using multi modal AI [160] provide opportunities for more efficient and extensive data collection. This could also improve the current aspects of the workflow. For example, by combining textual information about burial measurements, the plausibility of the assignment of specific arrows or scales to burials could be checked. Named entity recognition could also be used to extract information from the text that accompanies images. This has already been used to analyse field reports [45] and could automatically ascribe archaeological features and artefacts to sites, periods, and cultures.

Our software uses well established high performance neural networks that work well with most existing and even older hardware. There have been significant advances in two areas of machine learning: more performant hardware and better performing networks [161]. TPUs (tensor processing units) offer a significant improvement over conventional CPUs or GPUs for machine learning use cases. Newer network architectures, such as Vision Transformers [162–164] have greatly increased performance, compared to Resnet-152 or RetinaNet used in our application. Those improvements would reduce the number of manual corrections needed (**Table 4.3**).

An ethical risk of introducing *AutArch* to a wider audience is the potential automation of repetitive jobs by AI such as database entry and curation, typically performed by paid trainees and technicians. This risk is present in all sectors where AI is being deployed. However, *AutArch* requires a human agent to manually validate database entries. In theory, the AI-assisted workflow should make the recording process significantly less tedious, freeing up time for more creative tasks like literature search, data analysis and interpretation.

## Data Availability Statement

The data that support the findings of this study are openly available at [10.5281/zenodo.14999893](https://doi.org/10.5281/zenodo.14999893).

## Author Contributions

Conceptualization: K.K., M.B. Data curation: K.K., A.M., A.V.G. Formal analysis: K.K, A.M. Funding acquisition: M.B. Investigation: K.K, A.M., M.B, A.V.G. Methodology: K.K, A.W. Project administration: M.B. Resources: K.K., M.B., A.V.G. Software: K.K. Supervision: M.B. Validation: K.K., A.M., M.B., Y.D., V.H. R.L. Visualisation: K.K., A.M. Writing – original draft: K.K, A.M., M.B. Writing – review & editing: K.K, A.M., A.V.G., A.W., R.L., Y.D, M.B. All authors reviewed the manuscript.

## Acknowledgements

We thank Felix Riede for feedback on an earlier manuscript. Florian Helmecke, Katie Meheux, Louise Olerud, Stefano Palalidis and Bianca Preda-Bălănică kindly helped to identify relevant archaeological catalogues. We are also grateful to Miroslav Dobeš and Petr Limburský for the permission to reproduce illustrations from Vliněves. For help with copyright enquiries, we thank the NFDI4Culture legal helpdesk: Grischka Petri and Fabian Rack. We thank Benedikt Kirsch-Gerweck and Marin Althuis for their help with the installation guide. This research was supported by the German Research Foundation (DFG Project CO-MOVE, Grant n° 466680522, awarded to Maxime Brami). Volker Heyd's and Yoan Diekmann's contributions to this study were supported by the European Research Council (ERC) under the European Union's Horizon 2020 research and innovation program (Grant agreement n° 788616-YMPACT).

*Supplementary Materials for*

# An AI-assisted workflow for object detection and data collection from archaeological catalogues

***Authors***

Kevin Klein[1,2]*, Antoine Muller[3], Alyssa Wohde[4], Alexander V. Gorelik[1,2], Volker Heyd[5], Ralf Lämmel[6], Yoan Diekmann[1], Maxime Brami[1,2]*

***Affiliation***

[1]Palaeogenetics Group, Institute of Organismic and Molecular Evolution (iomE), Johannes Gutenberg University Mainz, Mainz, Germany.
[2]Vor- Und Frühgeschichtliche Archäologie, Institut Für Altertumswissenschaften, Johannes Gutenberg University Mainz, Mainz, Germany.
[3]SFF Centre for Early Sapiens Behaviour (SapienCE), University of Bergen, Bergen, Norway.
[4]University of Koblenz, Koblenz, Germany
[5]Department of Cultures / Archaeology, University of Helsinki, Helsinki, Finland
[6]Software Languages Team, University of Koblenz, Koblenz, Germany

* Correspondence to Kevin Klein (kkevin@students.uni-mainz.de) and Maxime Brami (mbrami@uni-mainz.de)

# SI 1. List of Online Resources

The GitHub repositories (https://github.com/kevin-klein/autarch-material, https://github.com/kevin-klein/autarch) contain the following assets:

- The complete AutArch source code with README, including installation instructions.
- The AI training datasets presented in Section 2.2 – due to copyright restrictions, we cannot include the underlying publications (or the images of graves therein or the objects therein such as skeletons and north arrows), but we provide a database dump of data extracted (detailing data summarised in the paper) and all labelling data used for training. Interested parties may use the original documents to train the same AI model or a different one, using comparable publications.
- The AI model presented in Section 2.1 – we provide a usable AI model that can be used to perform object detection and other steps of our workflow on the same (if available) or any suitable publications.
- The evaluation of object detection, workflow, and usability in Section 4 – the repository provides the raw data in the form of several spreadsheets used for the discussion in Section 4.

# SI 2. Supplementary Methods

## *General Object Detection Setup*

Our software detects objects by means of a 'deep learning' approach with artificial neural networks (ANNs). ANNs are mathematical function networks that can be thought of as layers of artificial neurons – also known as Perceptrons or Nodes. Neurons have one or several inputs with associated weights, a bias- and an activation function. For simple linear layers, each input is multiplied with its weight, the weighted inputs are summed, and the bias is added to it. The resulting value is the argument of the activation function. Evaluating the function gives the final result that is passed onto the next layer, which performs its calculations based on the output of the previous layer. Bias and input weights are the model parameters that need to be trained given annotated input data. For example, our software extensively uses Resnet-152 (Residual Neural Network), which has a total of 115.6 million parameters [1].

Optimisers are algorithms that are executed on the model and tweak its parameters to improve classification accuracy; increasing the number of times a correct label is produced for a given image. A loss function is used to quantify the prediction error, which is related to how often the model predicts incorrect labels. The output of an image classification network are the probabilities that an object belongs to each of the various labelled classes.

For classification, a widely used loss function [2] is cross entropy loss, defined as:

$$l(x, y) = -\sum_{c \in C} x(c) \log y(c),$$

where $C$ is the set of classes and $x, y$ the predicted and true probability vector, respectively. Training a model on the whole dataset once is referred to as an "epoch". Usually models are trained for hundreds or thousands of epochs. The training data is re-shuffled randomly before every training epoch to avoid learning effects caused by a specific order.

## *Converting PDF Pages to Images*

All documents used for this project were in PDF format. The uploaded documents were saved to a database and converted to separate images for each page. Vips::Image.pdfload is used to convert the files to single page images (ruby-vips 2.4.1 with libvips 8.9.1-2). Further PDF information such as the number of pages is obtained using pdf-reader 2.11.0. These images were then scaled and fed into an R-CNN, trained with a set of archaeological drawings.

## *Identifying Rotation using North Arrows*

The north arrows are analysed using a Resnet-152 [3] residual neural network that retrieves their angle. This network has been trained in an unsupervised manner as opposed to other neural networks discussed here. For every training step, those arrows are rotated by a random angle within 10° degree steps (0°, 10°, 20°, 30° …). The rotated image is then processed by the neural network, which predicts those 10° degree bins as a label. RMSProp [4] (root mean square propagation) was used as the optimizer with a learning rate of 1e-3. It was trained for 1000 epochs with a batch size of 16.

## *Contour Detection*

We use findContours from OpenCV 4.6 with RETR_EXTERNAL and CHAIN_APPROX_SIMPLE to detect the outline of the burial, the north arrow, the cross section and the scale. The page image was inverted, then converted to grayscale. After this step, threshold using the OpenCV threshold method with the parameters 40 (thresh), 255 (max) and THRESH_BINARY (type) was performed. Then the largest contour is selected using the length of its arc. This outline is used to calculate the area, width, length and rotation of the burial. The area is retrieved using the OpenCV contourArea function. Width and length are created by first obtaining the bounding rectangle using minAreaRect [5]. By definition, the width is the shorter side of the rectangle and length is the longer side. The angle is also derived from the bounding rectangle.

## *Skeleton Pose Classification Dataset*

The poses of skeletons are classified with Resnet-152. RMSProp was used as the optimizer with a learning rate of 1e-3. It was trained 1000 epochs with a batch size of 16. From the object detection training dataset, all skeleton images were extracted using their labelled bounding box. These images were manually assigned to two classes, “supine” and “flexed on the side”. All images that could not be clearly distinguished were removed. This resulted in 239 drawings of skeletons in supine position and 133 of skeletons flexed on one side. They were all resized to 300x300 pixels to provide a uniform resolution.

## *Arrow Classification Dataset*

All north arrows were extracted from the object detection training dataset. Then, a manual selection was performed to have an even spread of different types of arrows. In total, we chose 118 north arrows. These were scaled to 224x224 pixels and rotated to point towards 0° degrees.

## *Orientation of the Burials*

Two types of orientations can be used: a) skeletons within burials that have a discernible body position; or b) the orientation of the long side of the bounding box encapsulating the object. The outline orientation is given between 0 and 180 degrees, because the bounding box per se lacks direction.
In case an arrow through the spine of the skeleton was added, the angle is calculated as follows: $f(a_X, a_Y, a_\alpha, s) = atan^2(a_X, a_Y) - atan^2(-sin(a_\alpha)) - cos(-a_\alpha)$, with positive angles $a$ ($a = a' + 360°$ if $a' < 0°$) mapped to the interval $[0, 1]$ by the transformation $a\frac{180}{\pi}$, where $a_X$ and $a_Y$ are the x and y coordinate of the vector of the spine orientation arrow.

# SI 3. Training Dataset

The following resources, which include images of graves and grave contents, were used as training dataset for the object detection algorithm:

*Bader 1963* [6]*, Bader & Khalikov 1976* [7]*, Buchvaldek and Koutecký 1970* [8]*, Comsa 1989* [9]*, Dobeš & Limburský 2013* [10]*, Dumitrescu 1980* [11]*, Frînculeasa 2013* [12]*, Frînculeasa et al. 2015* [13]*, Frînculeasa et al. 2017* [14]*, Frînculeasa et al. 2018* [15]*, Frînculeasa 2019* [16]*, Frînculeasa 2020* [17]*, Frînculeasa et al. 2021* [18]*, Frînculeasa et al. 2022* [19]*, Gadzyackaya 1976* [20]*, Gymashian 1993* [21]*, Ivanova & Toshchev 2015 and references therein* [22]*, Kostyleva & Utkin 2010* [23]*, Krainov 1963* [24]*, Krainov 1964* [25]*, Langova 2009* [26]*, Limburský 2012* [27]*, Lougas et al. 2007* [28]*, Papac et al. 2021* [29]*, Rosetti 1959* [30]*, Toshchev, Shapovalov, Mikhailov 1991 and references therein* [31].

# SI 4. Evaluation by User Study

The test group consisted of undergraduate students, postgraduate and postdoctoral researchers from a variety of backgrounds, including archaeologists and non-archaeologists. The participants had no prior experience with the AI software presented in this manuscript. We also made sure that none of the participants were able to read Czech, because we wanted to prevent the participants from reading the textual descriptions instead of analysing the drawings [10].

An earlier version of our AI software, AutArch, however with the same AI models, was used in the user study. Based on the feedback and insight gathered during the study, we later made improvements to the software and to the graphical user interface. Participants were divided into two groups. The first group used Inkscape and then AutArch, while the second group used AutArch and then Inkscape in a typical AB/BA crossover study format. These two phases were timed and limited to 30 minutes.

The two groups were given a short induction to Inkscape. They were asked to import PDF pages containing graves from Dobeš and Limburský 2013 [10] into Inkscape, then manually measure each grave using the measuring function. A standard excel spreadsheet was provided with formulas to calculate the actual size of graves based on the scale. We also gave a short introduction on how to use AutArch. Participants were asked to record the grave ID, make necessary corrections to the bounding boxes, draw arrows through the spines and validate the scales, arrows and the deposition type. This publication[10] had already been imported into AutArch. We decided to have the participants import the PDF into Inkscape as opposed to it being imported already, because importing the whole PDF into Inkscape led to a significant performance reduction of the software. Participants were provided with a checklist for a number of steps – the current version of the software provides a more guided workflow instead of a checklist.

The results of the experiment are shown in **figure 7, Supplementary figures 1-2**. Manually collected data and the participants' AutArch results, i.e. width, length etc., were compared to a baseline that we created using AutArch ourselves and the percentage deviation was recorded. The deviation of all values for a single grave was summed and the average error was calculated between all graves for both Inkscape and AutArch. In instances where values were missing, these specific values were skipped in the error calculation. Situations where certain values could not be obtained, such as with missing cross-sections, were omitted. We observed that the use of our AI-software (AutArch) resulted in considerably fewer errors (in % of the baseline) and was faster than manual recording using Inkscape (number of graves for each participant in a 30-minute window).

Two of the participants exhibited larger errors than the other participants. We assume that different experience levels in working with computers affect the performance of our software at the same level as alternatives.

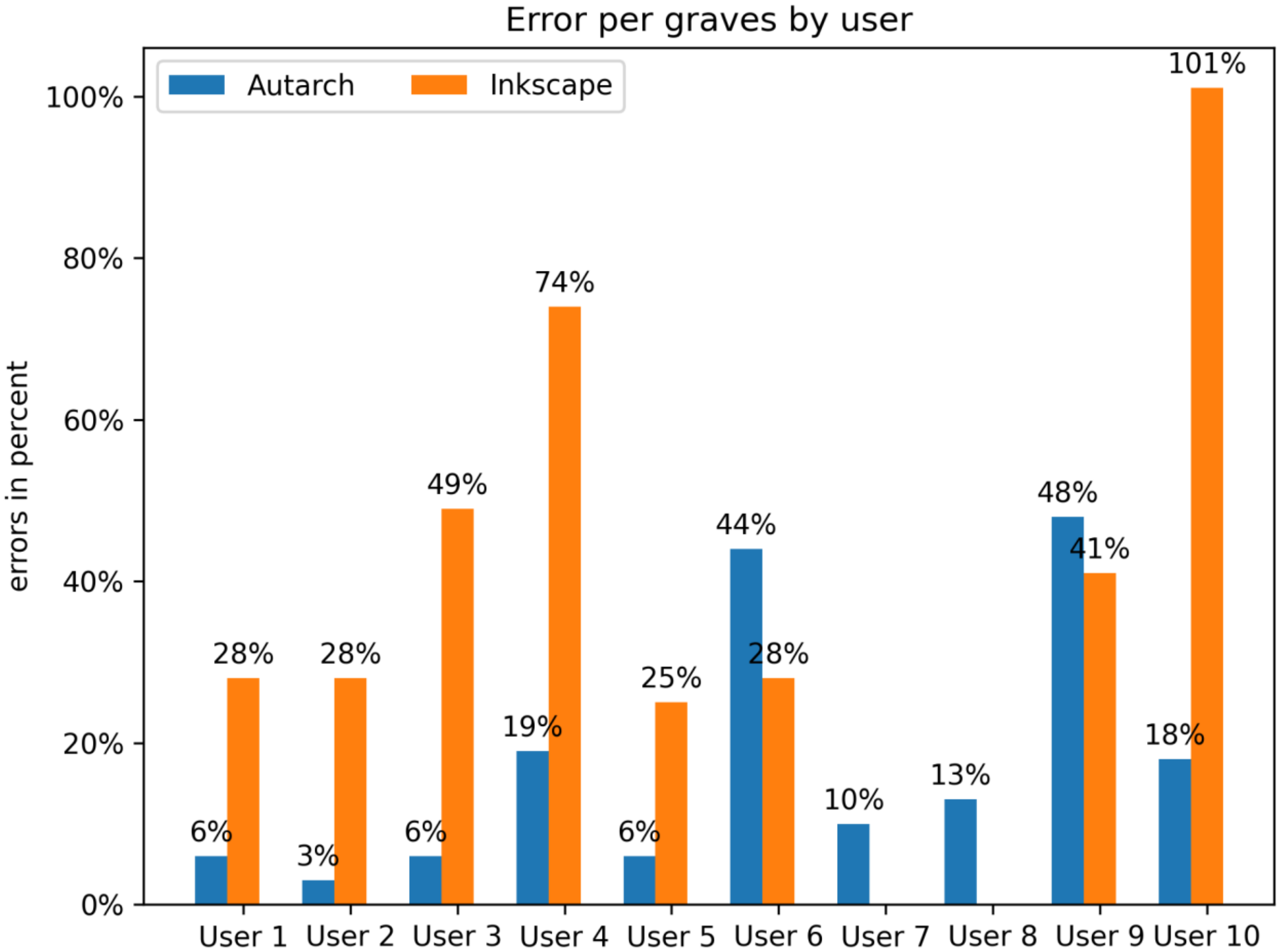

***Supplementary Figure 1.*** *Average errors for the dataset created by the user in the experiment relative to a baseline we created. No inkscape data was available for Users 7 and 8.*

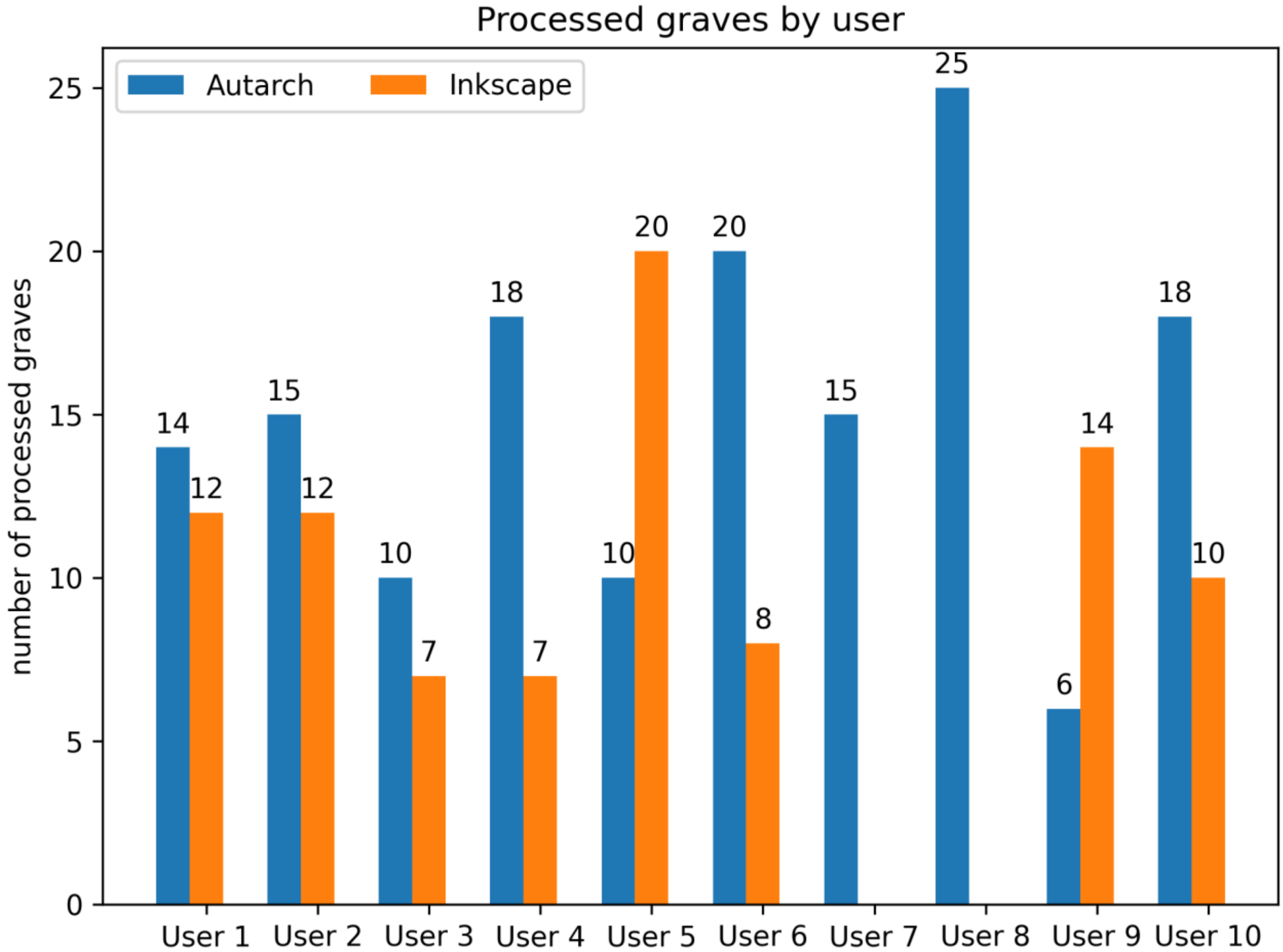

***Supplementary Figure 2.*** *Number of processed graves for the dataset created by the user in the experiment. No inkscape data was available for Users 7 and 8.*

# SI 5. Testing *AutArch* for Artefact Outline Extraction

To demonstrate the generalisability of *AutArch* to other archaeological materials, we conducted an analysis of arrowheads following the same outline extraction and EFA methods conducted for graves. AutArch was used to extract 553 complete arrowhead outlines from Nicolas'[32] comprehensive catalogue of sites from northwest France, southern Britain, and Denmark, spanning the Late Neolithic to the Early Bronze Age. A similar geometric analysis on this material was conducted by Matzig et al.[33], allowing us to test the potential of *AutArch* for conducting reproducible research. The mean shapes of the Late Neolithic, Bell Beaker, and Early Bronze arrowheads (**Supplementary Figure 3c**) are readily distinguishable, and the first two PCs of their EFA coefficients show much spatiotemporal patterning (**Supplementary Figure 3a**). We confirm the findings of Matzig et al.[33] that most variability is explained by barb morphology and edge curvature (PC1), as well as tang shape and arrowhead elongation (PC2).

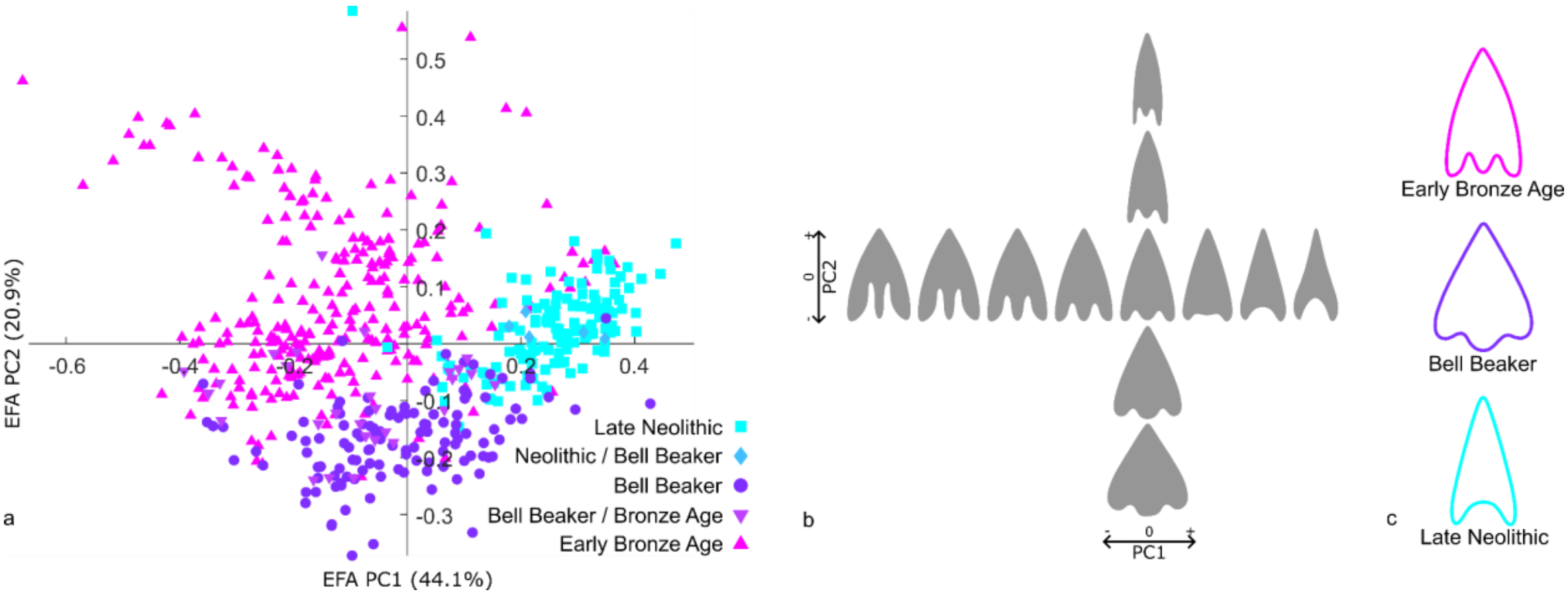

***Supplementary Figure 3.*** *Elliptical Fourier analysis of arrowheads. (a) PCA of the EFA coefficients of the sample of Neolithic, Bell Beaker, and Bronze Age arrowheads; (b) Reconstructed approximation of arrowhead shapes, plotted along the same axes as in d, in increments of 0.15; (c) Closed contours of the mean shapes of arrowheads in each period.*

# Supplementary References